\documentclass[11pt,a4paper]{article}
\usepackage[hyperref]{emnlp2018}
\usepackage{xcolor}
\usepackage{soul}
\usepackage[utf8]{inputenc}
\usepackage{CJK}
\usepackage{emnlp2018}
\usepackage{times}
\usepackage{url}
\usepackage{latexsym}
\usepackage{amsmath}
\usepackage{subcaption}
\usepackage{pgfplots}
\usepackage{pgfplotstable}
\pgfplotsset{width=7.0cm,compat=1.5}
\usepackage{color,xcolor,framed}

\aclfinalcopy 
\def\aclpaperid{1235}

\newcommand*{\affaddr}[1]{#1} 
\newcommand*{\affmark}[1][*]{\textsuperscript{#1}}
\newcommand*{\email}[1]{\texttt{#1}}

\title{Learning When to Concentrate or Divert Attention: \\Self-Adaptive Attention Temperature for Neural Machine Translation}

\author{Junyang Lin\affmark[1,2], Xu Sun\affmark[2], Xuancheng Ren\affmark[2], Muyu Li\affmark[2], Qi Su\affmark[1]\\
\affaddr{School of Foreign Languages, Peking University\affmark[1]}\\
\affaddr{MOE Key Lab of Computational Linguistics, School of EECS, Peking University\affmark[2]}\\
\email{\{linjunyang, xusun, renxc, limuyu0110, sukia\}@pku.edu.cn}\\
}
\date{}

\begin{document}
\maketitle
\begin{CJK}{UTF8}{gbsn}
\begin{abstract}
  Most of the Neural Machine Translation (NMT) models are based on the sequence-to-sequence (Seq2Seq) model with an encoder-decoder framework equipped with the attention mechanism. However, the conventional attention mechanism treats the decoding at each time step equally with the same matrix, which is problematic since the softness of the attention for different types of words (e.g. content words and function words) should differ. Therefore, we propose a new model with a mechanism called Self-Adaptive Control of Temperature (SACT) to control the softness of attention by means of an attention temperature. Experimental results on the Chinese-English translation and English-Vietnamese translation demonstrate that our model outperforms the baseline models, and the analysis and the case study show that our model can attend to the most relevant elements in the source-side contexts and generate the translation of high quality.
\end{abstract}

\section{Introduction}
\label{intro}

%
%

In recent years, Neural Machine Translation (NMT) has become the mainstream method of machine translation as it, in a great number of cases, outperforms most models based on Statistical Machine Translation (SMT), let alone the linguistics-based methods. One of the most popular baseline models is the sequence-to-sequence (Seq2Seq) model \citep{Kalchbrenner,seq2seq,ChoEA2014} with attention mechanism \citep{attention,stanfordattention}.
However, the conventional attention mechanism is problematic in real practice. The same weight matrix for attention is applied to all decoder outputs at all time steps, which, however, can cause inaccuracy. Take a typical example from the perspective of linguistics. Words can be categorized into two types, function word, and content word. Function words and content words execute different functions in the construction of a sentence, which is relevant to syntactic structure and semantic meaning respectively. Our motivation is that the attention mechanism for different types of words, especially function word and content word, should be different. When decoding a content word, the attention scores on the source-side contexts should be harder so that the decoding can be more focused on the concrete word that is semantic referent in the source text. But when decoding a function word, the attention scores should be softer so that the decoding can pay attention to its syntactic constituents in the source text that may be several words instead of one word.


To tackle the problem mentioned above, we propose a mechanism called Self-Adaptive Control of Temperature (SACT) to control the softness of attention for the RNN-based Seq2Seq model\footnote{The code is available at \url{https://github.com/lancopku/SACT}}. We set a temperature parameter, which can be learned by the model based on the attention in the previous decoding time steps as well as the output of the decoder at the current time step. With the temperature parameter, the model is able to automatically tune the degree of softness of the distribution of the attention scores. To be specific, the model can learn a soft distribution of attention which is more uniform for generating function word and a hard distribution which is sparser for generating content words. 

Our contributions in this study are in the following:
(1). We propose a new model for NMT, which contains a mechanism called Self-Adaptive Control of Temperature (SACT) to control the softness of the attention score distribution.
(2). Experimental results demonstrate that our model outperforms the attention-based Seq2Seq model in both Chinese-English and English-Vietnamese translation, with a 2.94 BLEU point and 2.19 BLEU score advantage respectively\footnote{What should be mentioned is that though the ``Transformer'' model is recently regarded as the best, the model architecture is not the focus of our study. Furthermore, our proposed mechanism can also be applied to the aforementioned model, which will be a part of our future study.}.
(3). The analysis shows that our model is more capable of translating long texts, compared with the baseline models.

\section{Our Model}

As is mentioned above, our model is substantially a Seq2Seq framework improved by the SACT mechanism. In this section, we first briefly describe the Seq2Seq model, then introduce the SACT mechanism in detail.

\subsection{Seq2Seq Model}
We implement the encoder with bidirectional Long Short-Term Memory (LSTM) \citep{LSTM}, where the encoder outputs from two directions at each time step are concatenated, and we implement the decoder with unidirectional LSTM. We train our model with the Cross-Entropy Loss, which is equivalent to the maximum likelihood estimation. In the following, we introduce the details of our proposed attention mechanism.

\subsection{Self-Adaptive Control of Temperature}

\begin{table*}[tb]
\setlength{\tabcolsep}{4.5pt}
\centering
    \begin{tabular}{l|c|c|c|c|c}
    \hline
    Model  & MT-03 & MT-04 & MT-05 & MT-06 & Ave.\\ \hline\hline
    Moses  & 32.43 & 34.14 & 31.47 & 30.81 & 32.21        \\ 
    RNNSearch  & 33.08 & 35.32 & 31.42 & 31.61 & 32.86        \\  
    Coverage & 34.49 & 38.34 & 34.91 & 34.25 & 35.49 \\
    MemDec & 36.16 & 39.81 & 35.91 & \textbf{35.98} & 36.97 \\
     \hline\hline
     Seq2Seq & 35.32  & 37.25 & 33.52  & 33.54  & 34.91 \\
    \textbf{+SACT}  & \textbf{38.16}  & \textbf{40.48} & \textbf{36.81}  & 35.95 & \textbf{37.85} \\
    \hline
    \end{tabular}
    \caption{\textbf{Results of the models on the Chinese-English translation}}
    \label{cnen}
\end{table*}


In our assumption, due to the various functions of words, decoding at each time step should not use the identical attention mechanism to extract the required information from the source-side contexts. Therefore, we propose our Self-Adaptive Control of Temperature (SACT) to improve the conventional attention mechanism, so that the model can learn to control the scale of the softness of attention for the decoding of different words. In the following, we present the details of our design of the mechanism.

We set a temperature parameter $\tau$ to control the softness of the attention at each time step. The temperature parameter $\tau$ can be learned by the model itself. In our assumption, the temperature parameter is learned based on the information of the decoding at the current time step as well as the attention in the previous time steps, referring to the information about what has been translated and what is going to be translated. Specifically, it is defined as below:
\begin{align}
\tau_{t} &= \lambda^{\beta_{t}} \\
\beta_t &= \tanh(W_{c}\tilde{c}_{t-1}+ U_{s}s_{t})
\end{align}
where $s_{t}$ is the output of the LSTM decoder as mentioned above, $\tilde{c}_{t-1}$ is the context vector generated by our attention mechanism at the last time step (initialized with the initial state of the decoder for the decoding at the first time step), and $\lambda$ is a hyper-parameter, which decides the upper bound and the lower bound of the scale for the softness of attention. To be specific, $\lambda$ should be a number larger than 1\footnote{In our experiments, we use $\lambda$ of different values, ranging from 2 to 10. The performance differences of models with different $\lambda$ values are not significant, and we report the results of the model with $4$ as the value of $\lambda$ as it achieves the best performance.}. The range of the output value of $\tanh$ function is $(-1, 1)$, so the range of the $\tau$ is $(\frac{1}{\lambda}, \lambda)$. Furthermore, the temperature parameter is applied to the conventional attention mechanism. 

Different from the conventional attention mechanism, the temperature parameter is applied to the computation of attention score $\alpha$ so that the scale of the softness of attention can be changed. We define the new attention score and context vector as $\tilde{\alpha}$ and $\tilde{c}$, which are computed as:
\begin{align}
\tilde{c}_{t} &= {\sum^{n}_{i=1} \tilde{\alpha}_{t,i}h_{i}} \\
{\tilde{\alpha}_{t,i}} &= \frac{{exp}(\tau^{-1}_{t}{e_{t,i}})}{{\sum_{j=1}^{n}}{exp}(\tau^{-1}_{t}{e_{t,j}})}
\end{align}
From the definition above, it can be inferred that when the temperature increases, the distribution of the attention score $\alpha$ is smoother, meaning that softer attention is required, and when the temperature is low, the distribution is sparser, meaning that harder attention is required. Therefore, the model can tune the softness of the attention distribution self-adaptively based on the current output for the decoder and the history of attention, and learns when to attend to only corresponding words and when to attend to more relevant words for further syntactic and semantic information.

\section{Experiment}
In the following, we introduce the experimental details, including the datasets and the experiment setting.

\subsection{Datasets}

\textbf{Chinese-English Translation} We train our model on 1.25M sentence pairs\footnote{The dataset is extracted from LDC2002E18, LDC2003E07, LDC2003E14, Hansards portion of LDC2004T07, LDC2004T08 and LDC2005T06} with 27.9M Chinese words and 34.5M English words, and we validate our model on the dataset for the NIST 2002 translation task and test our model on the datasets for the NIST 2003, 2004, 2005, 2006 translation tasks. We use the most frequent 30K words for the Chinese vocabulary and the English vocabulary respectively, covering about 97.4\% and 99.7\% of the corpora. The evaluation metric is case-insensitive BLEU score computed by \texttt{mteval-13a.perl} \citep{bleu}.

\noindent \textbf{English-Vietnamese Translation} The training data is from the translated TED talks, containing 133K training sentence pairs provided by the IWSLT 2015 Evaluation Campaign \citep{2015iwslt}. The validation set is the TED tst2012 with 1553 sentences and the test set is the TED tst2013 with 1268 sentences. The English vocabulary is 17.7K words and the Vietnamese vocabulary is 7K words. The evaluation metric is also BLEU as mentioned above\footnote{For comparison with the existing system, we use \texttt{multi-bleu.perl} instead.}.

\subsection{Setting}
Our model is implemented with PyTorch on an NVIDIA 1080Ti GPU. Both the size of word embedding and the size of the hidden layers in the encoder and decoder are 512. Gradient clipping for the gradients is applied with the largest gradient norm 10 in our experiments. Dropout is used with the dropout rate set to 0.3 for the Chinese-English translation and 0.4 for the English-Vietnamese translation, in accordance with the evaluation on the development set. Batch size is set to 64. We use Adam optimizer \citep{KingmaBa2014} to train the model\footnote{$\alpha=0.0003$, $\beta_{1}=0.9$, $\beta_{2}=0.999$ and $\epsilon=1\times10^{-8}$}.

\subsection{Baselines}
In the following, we introduce our baseline models for the Chinese-English translation and the English-Vietnamese translation respectively.

For the Chinese-English translation, we compare our model with the most recent NMT systems, illustrated in the following. \textbf{Moses} is an open source phrase-based translation system with default configurations and a 4-gram language model trained on the training data for the target language; \textbf{RNNSearch} is an attention-based Seq2Seq with fine-tuned hyperparameters; \textbf{Coverage} is the attention-based Seq2Seq model with a coverage model \citep{coverage}; \textbf{MemDec} is the attention-based Seq2Seq model with the external memory \citep{memdec}.

For the English-Vietnamese translation, the models to be compared are presented below. \textbf{RNNSearch} The attention-based Seq2Seq model as mentioned above, and we present the results of \citep{luong2015stanford}; \textbf{NPMT} is the Neural Phrase-based Machine Translation model by \citet{nplm}.

\section{Results and Analysis}

In the following, we present the experimental results as well as our analysis of temperature and case study.

\subsection{Results}
We present the performance of the baseline models and our model on the Chinese-English translation in Table \ref{cnen}. As to the recent models on the same task with the same training data, we extract their results from their original articles. Compared with the baseline models, our model with the SACT for the softness of attention achieves better performance, with the advantages of BLEU score 2.94 over the conventional attention-based Seq2Seq model. The SACT effectively learns the temperature to control the softness of attention so that the model can utilize the information from the source-side contexts more efficiently. 

We present the results of the models on the English-Vietnamese translation in Table \ref{envi}. Compared with the attention-based Seq2Seq model, our model with the SACT can outperform it with a clear advantage of 2.17 BLEU score. We also display the most recent model NPMT \citep{nplm} trained and tested on the dataset. Compared with NPMT, our model has an advantage of BLEU score of 1.43. It can be indicated that for low-resource translation, the information from the deconvolution-based decoder is important, which brings significant improvement to the conventional attention-based Seq2Seq model.

\begin{table}[tb]
\centering
    \begin{tabular}{l|c}
    \hline
    Model & BLEU  \\ \hline\hline
    RNNSearch &     26.10                 \\ 
    NPMT &   27.69               \\
    \hline\hline
     Seq2Seq &  26.93 \\
    \textbf{+SACT} &  \textbf{29.12} \\
    \hline
    \end{tabular}
    \caption{\textbf{Results of the models on the English-Vietnamese translation} }
    \label{envi}
\end{table}


\subsection{Analysis}

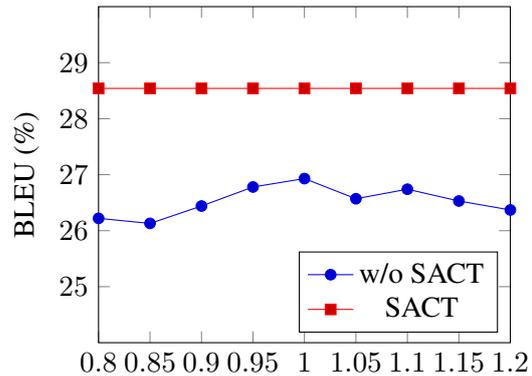
\begin{figure}[tb]
\centering



\begin{tikzpicture}
\selectcolormodel{RGB}
\begin{axis}[legend pos=south east, xlabel={}, ylabel = {BLEU (\%)}, xmin=0.80, xmax=1.20, ymin=24.0, ymax=30.0, xtick=data, ytick={25.0,26.0,27.0,28.0,29.0}]
\addlegendentry{w/o SACT}
\addplot 
coordinates{(0.80, 26.22)(0.85, 26.13)(0.90, 26.44)(0.95, 26.78)(1.00, 26.93)(1.05, 26.57)(1.10, 26.74)(1.15, 26.53)(1.20, 26.37)};

\addlegendentry{SACT}
\addplot 
coordinates{(0.80, 28.54)(0.85, 28.54)(0.90, 28.54)(0.95, 28.54)(1.00, 28.54)(1.05, 28.54)(1.10, 28.54)(1.15, 28.54)(1.20, 28.54)};
\end{axis}
\end{tikzpicture}
\caption{BLEU scores of the Seq2Seq models with fixed values for the temperature parameter. Models are tested on the test set of the English-Vietnamese translation.}
\label{lambda}

\end{figure}

In order to verify whether the automatically changing temperature can positively impact the performance of the model, we implement a series of models with fixed values, ranging from $0.8$ to $1.2$, for the temperature parameter. From the results shown in Figure\ref{lambda}, it can be found that the automatically changing temperature can encourage the model to outperform those with fixed temperature parameter.

\begin{figure}
\centering
\subcaptionbox{}[.45\textwidth]{
\centering
\includegraphics[height=2.0cm]{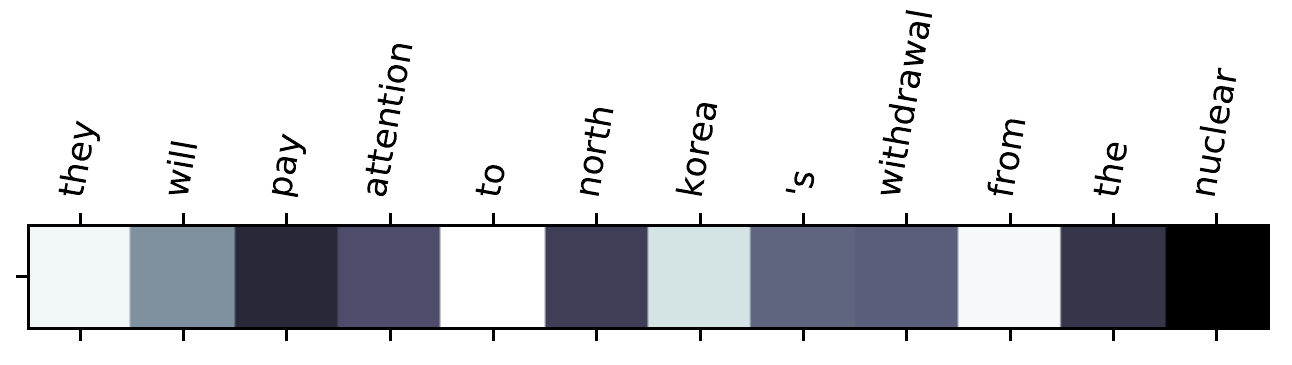}
}
\hspace{0.5in}
\subcaptionbox{}[.45\textwidth]{
\centering
\includegraphics[height=1.88cm]{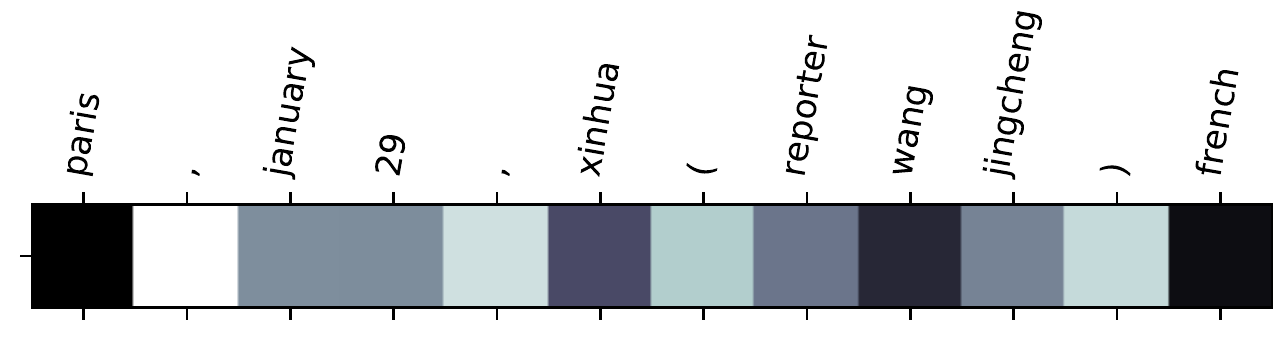}
}
\caption{\textbf{Examples of the heatmaps of temperature parameter} The dark color refers to low temperature, while the light color refers to high temperature.}
\label{temperature}
\end{figure}

Furthermore, as our model generates a temperature parameter at each time step of decoding, we present the heatmaps of two translations from the testing on the NIST 2003 for the Chinese-English translation on Figure \ref{temperature}. From the heatmaps, it can be found that the model can adapt the temperature parameter to the generation at the current time step. In Figure \ref{temperature}(a), when translating words such as ``to''  and ``from'', which are syntactic-relevant prepositions and both lack direct corresponding words in the source text or pronoun such as ``they'', whose corresponding word ``tamen'' in the source may be a part of the possessive case or the objective case, the temperature parameter increases to soften the attention distribution so that the model can attend to more relevant elements for accurate extraction of the information from the source-side contexts. On the contrary, when translating content words or phrases such as ``pay attention'' and  ``nuclear'', where there are direct corresponding words ``zhuyi'' and ``hezi'' in the source text, the temperature decreases to harden the attention distribution so that the model can focus on the corresponding information in the source text for accurate translation. In Figure \ref{temperature}(b), the temperature parameters for the punctuations are high as they are highly connected to the syntactic structure and those for the content words with concrete correspondences such as location ``paris'', name of organization ``xinhua'', name of person ``wang'' and nationality ``french''.

\subsection{Case Study}
We present two examples of the translation of our model in comparison with the translation of the conventional attention-based Seq2Seq model and the golden translation. In Table \ref{example}(a), it can be found that the translation of the conventional Seq2Seq model does not give enough credit to the word ``chengzhang'' (meaning ``growth''), while our model can not only concentrate on the word but also recognize the word as a noun (``chengzhang'' in Chinese can be both noun and verb). Even compared with the golden translation, the translation of our model seems better, which is a grammatical and coherent sentence. In Table \ref{example}(b), although the Seq2Seq model can generate the translation about the increase in the crude oil, it wrongly connects the increase with the threat of war in Iraq. In contrast, as our model has more capability of analyzing the syntactic structure by softening the attention distribution in the generation of syntax-relevant words, it extracts the causal relationship in the source text and generates the correct translation.

\begin{table}[tb]
\small
\setlength{\tabcolsep}{3pt}
\centering
    \subcaptionbox{}{
    \begin{tabular}{p{7.5cm}}
    \hline
    \textbf{Source:} 中国 大陆 手机 用户 成长 将 减缓\\ 
    \hline
    \textbf{Gold:} growth of mobile phone users in mainland china to slow down\\
    \hline  
    \textbf{Seq2Seq:} mainland cell phone users slow down\\
    \hline
    \textbf{SACT:} \colorbox{yellow}{the growth of} cell phone users \colorbox{yellow}{in chinese} mainland \colorbox{yellow}{will} slow down
\\
    \hline
    \end{tabular}}
    \subcaptionbox{}{
    \begin{tabular}{p{7.5cm}}
    \hline
    \textbf{Source:} 自 去年 12 以来, 受 委内瑞拉 国内 大罢工 和 伊拉克 战争 的影响, 国际 市场 原油 价格 持续 上涨 。\\ 
    \hline
    \textbf{Gold:} since december last year , the price of crude oil on the international market has kept rising due to the general strike in venezuela and the threat of war in iraq .\\
    \hline  
    \textbf{Seq2Seq:} since december last year , the international market has continued to rise \colorbox[rgb]{0.99,0.86,0.86}{in the international market} and the threat of the iraqi war \colorbox[rgb]{0.99,0.86,0.86}{has continued to rise} .\\
    \hline
    \textbf{SACT:} since december last year , the international market of crude oil has continued to rise \colorbox{yellow}{because of the strike in venezuela and the war in iraq}.
\\
    \hline
    \end{tabular}}
    \caption{Two examples of the translation on the NIST 2003 Chinese-English translation task. The difference between Seq2Seq and SACT is shown in color.}
    \label{example}
\end{table}

\section{Related Work}
Most systems for Neural Machine Translation are based on the sequence-to-sequence model (Seq2Seq) \citep{seq2seq}, which is an encoder-decoder framework \citep{Kalchbrenner,ChoEA2014,seq2seq}. To improve NMT, a significant mechanism for the Seq2Seq model is the attention mechanism \citep{attention}. Two types of attention are the most common, which are proposed by \citet{attention} and \citet{stanfordattention} respectively. 

Though the attention mechanism is powerful for the requirements of alignment in NMT, some prominent problems still exist. To tackle the impact of the attention history\citet{coverage,micover, interactive, memdec,decoding-history-based} take the attention history into consideration. An important breakthrough in NMT is that \citet{googleattention} applied the fully-attention-based model to NMT and achieved the state-of-the-art performance. To further evaluate the effect of our attention temperature mechanism, we will implement it to the ``Transformer'' model in the future. Besides, the studies on the attention mechanism have also contributed to some other tasks \citep{global_encoding,table2text}

Beyond the attention mechanism, there are also important methods for the Seq2Seq that contribute to the improvement of NMT. \citet{BOW} incorporates the information about the bag-of-words of the target for adapting to multiple translations, and \citet{global_decoding} takes the target context into consideration. 

\section{Conclusion and Future Work}
\label{conclusion}
In this paper, we propose a novel mechanism for the control over the scope of attention so that the softness of the attention distribution can be changed adaptively. Experimental results demonstrate that the model outperforms the baseline models, and the analysis shows that our temperature parameter can change automatically when decoding diverse words. In the future, we hope to find out more patterns and generalized rules to explain the model's learning of the temperature.


\section*{Acknowledgements}
This work was supported in part by National Natural Science Foundation of China (No. 61673028) and the National Thousand Young Talents Program. Qi Su is the corresponding author of this paper.

\end{CJK}
\bibliography{emnlp2018}
\bibliographystyle{acl_natbib_nourl}

\end{document}